\documentclass[letterpaper, 10 pt, conference]{ieeeconf} 
\IEEEoverridecommandlockouts                             
\usepackage{color}
\usepackage{graphicx} 
\usepackage{mathtools} 
\usepackage{siunitx}
\usepackage{multirow}
\usepackage{float}
\usepackage{aliascnt}
\newaliascnt{eqfloat}{equation}
\newfloat{eqfloat}{h}{eqflts}
\floatname{eqfloat}{Equation}
\usepackage{array}
\usepackage{stfloats}

\newcommand*{\ORGeqfloat}{}
\let\ORGeqfloat\eqfloat
\def\eqfloat{%
  \let\ORIGINALcaption\caption
  \def\caption{%
    \addtocounter{equation}{-1}%
    \ORIGINALcaption
  }%
  \ORGeqfloat
}

\usepackage{etoolbox}
\newbool{showcolor}
\setbool{showcolor}{false}
\newcommand{\changed}[1]{%
  \ifbool{showcolor}{\textcolor{blue}{#1}}{#1}%
}

\title{\LARGE \bf Evaluating Accuracy of Vine Robot Shape Sensing\\ with Distributed Inertial Measurement Units}

\author{Alexis E. Laudenslager$^{1}$, Antonio Alvarez Valdivia$^{2}$, Nathaniel Hanson$^{2}$, and Margaret McGuinness$^{1}$
\thanks{$^{1}$Department of Aerospace and Mechanical Engineering, University of Notre Dame, Notre Dame, Indiana, USA.}
\thanks{$^{2}$Lincoln Laboratory, Massachusetts Institute of Technology, Lexington, Massachusetts, USA}
\thanks{Correspondence: \tt\small mmcguinness@nd.edu}%
\thanks{DISTRIBUTION STATEMENT A. Approved for public release. Distribution is unlimited.
This material is based upon work supported by the Department of the Air Force under Air Force Contract No. FA8702-15-D-0001 or FA8702-25-D-B002. Any opinions, findings, conclusions or recommendations expressed in this material are those of the author(s) and do not necessarily reflect the views of the Department of the Air Force.
© 2026 Massachusetts Institute of Technology.
Delivered to the U.S. Government with Unlimited Rights, as defined in DFARS Part 252.227-7013 or 7014 (Feb 2014). Notwithstanding any copyright notice, U.S. Government rights in this work are defined by DFARS 252.227-7013 or DFARS 252.227-7014 as detailed above. Use of this work other than as specifically authorized by the U.S. Government may violate any copyrights that exist in this work.}}

\begin{document}

\maketitle
\thispagestyle{empty}
\pagestyle{empty}

\begin{abstract}


\changed{Soft, tip-extending vine robots are well suited for navigating tight, debris-filled environments, making them ideal for urban search and rescue. Sensing the full shape of a vine robot's body is helpful both for localizing information from other sensors placed along the robot body and for determining the robot's configuration within the space being explored.
Prior approaches have localized vine robot tips using a single inertial measurement unit (IMU) combined with force sensing or length estimation, while one method demonstrated full-body shape sensing using distributed IMUs on a passively steered robot in controlled maze environments. However, the accuracy of distributed IMU-based shape sensing under active steering, varying robot lengths, and different sensor spacings has not been systematically quantified.
In this work, we experimentally evaluate the accuracy of vine robot shape sensing using distributed IMUs along the robot body. We quantify IMU drift, measuring an average orientation drift rate of 1.33$^{\circ}$/min across 15 sensors. For passive steering, mean tip position error was 11\% of robot length. For active steering, mean tip position error increased to 16\%. During growth experiments across lengths from 30–175 cm, mean tip error was 8\%, with a positive trend with increasing length. We also analyze the influence of sensor spacing and observe that intermediate spacings can minimize error for single-curvature shapes. These results demonstrate the feasibility of distributed IMU-based shape sensing for vine robots while highlighting key limitations and opportunities for improved modeling and algorithmic integration for field deployment.}


\end{abstract}

\section{Introduction}

In the event of a building collapse, survivors may exist in void spaces until rescued by urban search and rescue teams~\cite{collapsebasics}. Collapse zones contain electrical, chemical, and mechanical hazards that make it difficult for humans to find and rescue survivors~\cite{osha}, and void spaces are often too small for humans to enter. 
In one past disaster, 
a snake robot provided underground camera footage to rescuers~\cite{snake}, showing the promise of continuum robots for usefully navigating the confined spaces found in collapsed buildings.

Vine robots~\cite{al2025tip} are continuum soft robots that ``grow" by everting their tip using internal fluid pressure, similar to how a plant grows outward from the tip while its base remains stationary. These robots are well suited for cluttered environments, such as rubble sites, due to their ability to move without disturbing rubble~\cite{mishima2006development}, grow through spaces smaller than their diameter~\cite{hawkes2017soft}, and flex around curves~\cite{ataka2020model}. 

Two vine robots have been tested in urban search and rescue training sites~\cite{RoBoa,sprout}, each carrying a camera at their tip. While both showed promising results, neither implemented a method for sensing the full shape of the robot, which is crucial for localizing information from sensors placed along the robot body (e.g., to measure temperature and humidity~\cite{charm} or other quantities of interest for a rescue team). Full body shape sensing is also critical for determining the robot's configuration within the space being explored, to provide the human operator with situational awareness when the robot is outside their direct line of sight~\cite{vinebasic}.

\begin{figure}
    \centering
    \includegraphics[width=\columnwidth]{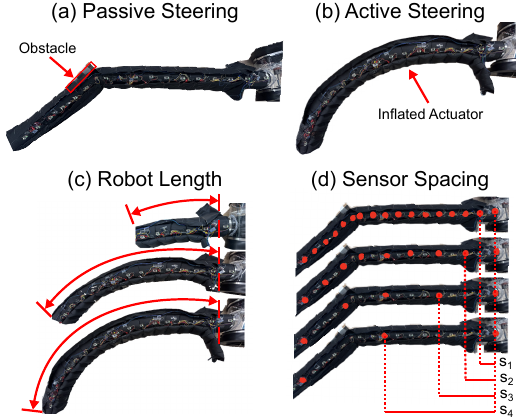}
    \vspace{-2.0em}
    \caption{\textbf{Summary of contributions.} We quantified the shape sensing accuracy of our vine robot instrumented with inertial measurement unit (IMU) sensors along its length at various \changed{(a)} passive and \changed{(b)} active steering angles, \changed{(c)} robot lengths, and \changed{(d)} sensor spacings. We also measured the effect on individual IMU orientation accuracy of 
    drift over time.}
    \label{fig:glamour}
    \vspace{-2.0em}
\end{figure}


One method for sensing the full shape of a vine robot's body~\cite{charm} uses inertial measurement units (IMUs) placed along the outer wall of the robot's body, combined with a model of how vine robots bend in contact with obstacles in the environment. This method was demonstrated on a passively steered vine robot (i.e., one that simply grows along the obstacles placed in front of it) in two laboratory mazes. 

However, the vine robots tested for urban search and rescue have operated using active steering (either with an internal articulated device~\cite{RoBoa} or with series pouch motor actuators~\cite{sprout} placed along the robot body's length). 
Knowing how the accuracy of vine robot shape sensing changes 
as the sensors have drifted with time, at various steering angles for both passive and active steering, at various robot lengths, and at various sensor spacings is critical for making sense of data obtained in the field.


The contributions of this work (also shown in Fig.~\ref{fig:glamour}) are:
\begin{itemize}
  \item Implementing vine robot shape sensing using distributed IMUs on an actively steered vine robot;
  \item Testing the effect of 
  drift over time on the orientation error of individual IMUs; and
  \item Quantifying the effect of steering angle, robot length, and sensor spacing on shape sensing accuracy.
\end{itemize}

\section{Related Work}

For other continuum robots besides vine robots, IMUs have shown great shape sensing promise. In~\cite{tape_def, IMU_shape_2, IMU_along, sheet_shape}, and~\cite{shaevitz2025shape}, IMUs were placed along a variety of flexible materials and continuum robots, and produced effective shape measurements, some providing tip position accuracy as high as within 1-3\% of the robot's length. Most algorithms involve modeling the shape of the structure between IMUs using ideas from Euler-Bernoulli beam bending and/or constant curvature models, which do not apply directly to vine robots due to their hollow internal structure.

For vine robots, three previous works have implemented shape sensing using IMUs. First is the previously discussed work~\cite{charm} that our work builds upon, which placed distributed IMUs along the body of a passively steered vine robot, allowing full body shape sensing with a tip localization accuracy of approximately 9\% of the robot's length. Another work~\cite{frias} used an IMU attached to a cap at the robot tip, in combination with force sensing at the robot tip, to estimate a passively steered robot's tip location based on the measured contact force and angle with the environment, as well as known patterns of how vine robots grow along obstacles. A third work~\cite{qin20253d} used an IMU attached to a cap at the robot tip, in combination with an encoder on the motor at the robot base, which allowed estimation of the robot's length, to determine the robot's tip location. However, the latter two approaches do not allow localization of the full robot body.

\section{Test Setup Architecture} 
\label{testarch}

\begin{figure}[t]
    \centering
    \includegraphics[width=\columnwidth]{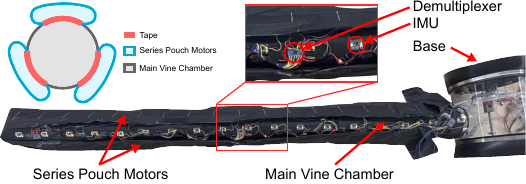}
    \vspace{-2.0em}
    \caption{\textbf{Vine robot shape sensing system.} 
    \changed{The vine robot includes a main chamber, three series pouch motor (SPM) steering actuators, and a base for growth and retraction. Eighteen IMUs are distributed along the robot body and connected through multiplexers to a microcontroller at the base.}
    }
    \label{fig:test_setup}
    \vspace{-0.5em}
\end{figure}

To implement distributed IMU-based shape sensing on an actively steered vine robot, we used a robot body made up of a main vine chamber, along with three series pouch motor actuators~\cite{vinebasic} for steering. \changed{Fig.~\ref{fig:test_setup} shows the vine robot shape sensing architecture, including the main body, steering actuators, distributed IMUs, and base system.}
Both the main chamber and actuators were constructed out of thermoplastic polyurethane-coated ripstop nylon fabric. Our setup also included a base~\cite{vinebasic} to control growth and retraction and a layered multiplexer-sensor setup to collect IMU orientation data. The vine robot had a flat diameter of 20.3 cm and inflated diameter of 12.9 cm, the minimum diameter that allowed for easy growth when wired with off-the-shelf Adafruit BNO055 sensor units. The system used 18 BNO055 IMUs spaced 10.2 cm apart along the top side of the robot (between two series pouch motor actuators), which allowed for 1.73 meters of robot length for shape estimation testing. 
We used a layered system of multiplexers (TCA9548A I2C) to collect data and identify the source from all 18 IMUs. 

\begin{figure}[t]
    \centering
    \includegraphics[width=0.9\columnwidth]{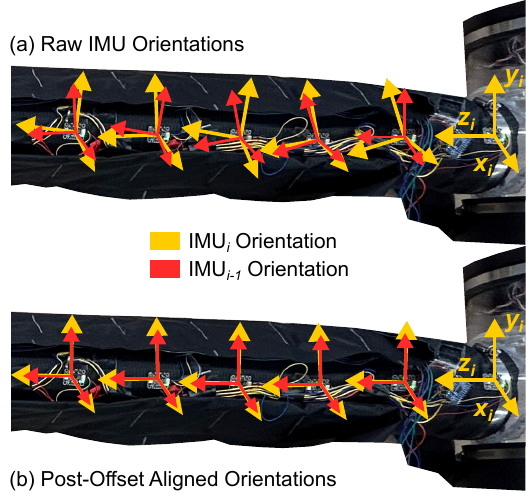}
    \vspace{-1.0em}
    \caption{\textbf{Orientation offsetting procedure.} 
    \changed{Yellow reference frames represent the orientation of each \changed{IMU$_i$}, and red reference frames represent the orientation of \changed{IMU$_{i-1}$}, the previous IMU (closer to the base). (a) Raw IMU orientations when the robot is straight. (b) After offsetting, consecutive IMU orientations are aligned by subtracting the stored relative orientation measured in the straight configuration.}
    }
    \label{fig:ref_frame}
    \vspace{-1.0em}
\end{figure}

Before collecting data, the IMUs were calibrated. Since the IMUs were secured to the vine robot, it was more efficient to calibrate them as a group rather than individually. First, we retracted \changed{the robot with} all IMUs into the base. Next, per the BNO055 datasheet~\cite{BNO055DS} procedure, we placed the base in several stationary configurations, then \changed{lifted the base to rotate and translate it} until all IMUs reported the highest level of calibration. 

Because each IMU unit measures its own orientation relative to a global reference frame slightly differently, and the IMUs may be placed differently on the robot body, an ``offsetting'' procedure was used to align the IMU reference frames before collecting data, as shown in Fig.~\ref{fig:ref_frame}. After calibration, the robot was grown to its full length and inflated so that the robot was in a straight configuration. Initial IMU orientation data was collected, and the relative orientation between consecutive IMUs was calculated by computing the orientation quaternion for each IMU, and then multiplying consecutive quaternions to find their 3D offset in the reference frame of the first IMU. These offset values were then removed from the raw orientation data at all later timesteps to align IMU reference frames throughout the duration of the experiment. \changed{After offsetting, each IMU measurement represents the relative rotation between adjacent robot segments, rather than the absolute orientation in the global frame.}

\section{Shape Sensing Algorithm}
\label{sec:algorithm}

\begin{figure}
    \centering
    \includegraphics[width=0.95\columnwidth]{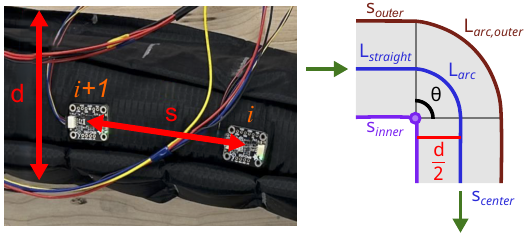}
    \vspace{-1.0em}
    \caption{\textbf{Vine robot model parameters used for shape sensing algorithm.}
    \changed{Consecutive IMUs, indexed $i$ and $i+1$, define a local segment of length $s$} with inflated diameter $d$. An orientation difference $\theta$ between IMUs is modeled as a single bend located midway between them, consisting of a circular arc of length $L_{arc}$ and two straight segments of length $L_{straight}$.}
    \label{fig:model}
    \vspace{-1.5em}
\end{figure}

The shape sensing algorithm used in this paper is identical to that used in \cite{charm}; the relevant parameters are shown in Fig.~\ref{fig:model}. \changed{Consecutive IMUs, indexed $i$ and $i+1$, define a local segment of the robot of length $s$ when the robot is straight.  The model depends on three primary variables:~the straight spacing $s$ between consecutive IMUs, the inflated robot diameter $d$, and the measured orientation of each IMU. The orientation difference between consecutive IMUs is computed in axis-angle notation, where $\theta$ represents the rotation angle between IMU frames.}
The model assumes that the robot will bend at a hinge point located halfway between \changed{IMU$_i$ and IMU$_{i+1}$}, and that the vine robot body material is inextensible, such that the outer edge of the robot's body does not extend during bending. Because of this assumption, the total outer-edge length $s_{outer}$ remains equal to the straight spacing $s$, while the centerline and inner edge shorten as the robot bends.
Next, the angle $\theta$ determines the necessary circular arc length $L_{arc,outer}$ to bend the outer edge of the robot, and the corresponding arc length $L_{arc}$ to bend the centerline of the robot. Finally, the arc length of the outer edge is subtracted from the total length of the outer edge $s_{outer}$, and the remaining length is split evenly as straight segments of length $L_{straight}$ between the bend and each IMU. The section of the length taken up by the arc is placed exactly halfway between the sensors. Equations \ref{eq1}-\ref{eq4} demonstrate the relationships between the parameters and allow us to compute $L_{arc}$ and $L_{straight}$, our desired values for our model. 

\begin{equation}
    \label{eq1}
    s_{outer} = s
\end{equation}
\begin{equation}
    \label{eq2}
    L_{arc,outer} = \theta*d
\end{equation}
\begin{equation}
    \label{eq3}
    L_{arc} = \frac{\theta*d}{2}
\end{equation}
\begin{equation}
    \label{eq4}
    L_{straight} = \frac{s-L_{arc,outer}}{2}
\end{equation}

Beginning at the first pair of IMUs, we used Equations \ref{eq1}-\ref{eq4} and the orientation of each IMU to find the end of the shape segment between \changed{IMU$_i$ and IMU$_{i+1}$. The end point and direction of this segment become} the start point and direction of the next shape segment\changed{, defined by IMU$_{i+1}$ and IMU$_{i+2}$}. This process is repeated until the full shape is constructed. 

\section{\changed{Drift} Experiments and Results}

\label{sec:drift}

\begin{figure}[t]
    \centering
    \includegraphics[width=\columnwidth]{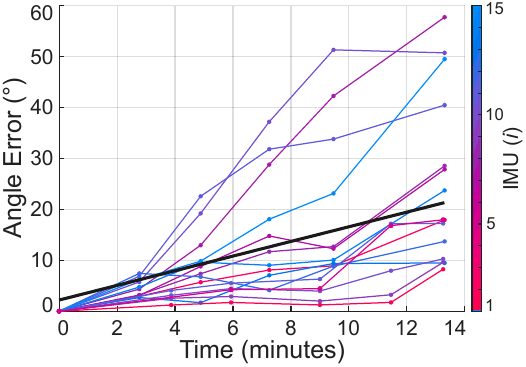}
    \vspace{-2.0em}
    \caption{\textbf{IMU drift test results.} 
    \changed{Orientation error relative to the initial offset shown as a function of time since offsetting. Each color corresponds to one of 15 IMUs, with one representative trial per unit. The black line indicates a linear fit across all data ($1.33^{\circ}$/min, $R^2$ = 0.16, $p$=0.0024).}}
    \label{fig:drift_result_new}
    \vspace{-1.50em}
\end{figure}

This experiment (Fig.~\ref{fig:drift_result_new}) quantifies IMU orientation error as a function of time due to drifting. We collected data by reprocessing the data for the Passive Steering Test and Active Steering Test discussed in the next section and comparing the IMU offset values recorded at various times. 

Fig.~\ref{fig:drift_result_new} demonstrates the drift experienced by 15 IMU units over time. \changed{Each line represents one representative trial for a single IMU. IMU$_1$ is located closest to the base, and IMU$_{15}$ is located nearest the tip of the robot. Drift magnitude varies across units, indicating sensor-dependent behavior. There is a statistically significant positive correlation between IMU orientation error and time since initial offset (linear fit with a slope of $1.33^{\circ}/\text{min}$, $R^2$ = 0.16, $p = 0.0024$).}

\changed{As described in Section~\ref{testarch}, to mitigate drift during the passive and active steering experiments, an offsetting procedure was performed immediately prior to each trial. Across all experiments, the average time between offsetting and data collection was less than 90 seconds. Based on the measured drift rate ($1.33^{\circ}/\text{min}$), this corresponds to less than approximately $2^{\circ}$ of expected drift-induced error during any individual trial.}

\section{Vine-Based Experiments and Results}

\subsection{Passive Steering Test}

\begin{figure*}[t]
    \centering
    \includegraphics[width=\textwidth]{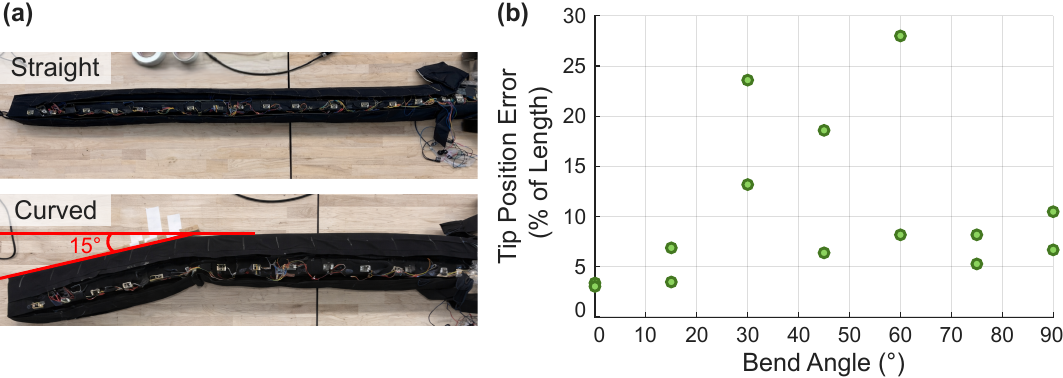}
    \vspace{-2.0em}
    \caption{\textbf{Passive steering test setup and results.} (a) At full length (1.73 m), the vine robot was bent \changed{against a wooden plank to prescribed} angles between $0^{\circ}$ and $90^{\circ}$ in $15^{\circ}$ increments. \changed{Two} trials were conducted at each angle. 
    \changed{(b) Tip position error as a percentage of the robot length, computed using an overhead photo as ground truth, plotted versus bend angle. No statistically significant relationship between bend angle and tip position error was observed ($p = 0.8$).} On average, the error is approximately 11\% of the robot length.}
    \label{fig:passive_setup_results}
    \vspace{-0.50em}
\end{figure*}

In this experiment (Figure \ref{fig:passive_setup_results}), a wooden plank was positioned at varying angles relative to the vine robot’s path. As the robot \changed{contacted} the obstacle, it bent accordingly, and the resulting ground truth and tip position estimates were recorded. This allowed us to analyze how bend angle affected the error in estimated tip position. Angles \changed{of the plank} were adjusted from $0^{\circ}$ to $90^{\circ}$ in $15^{\circ}$ increments, and the robot's length was kept constant. Offsetting was completed between each bend angle to minimize the effects of drift. Figure \ref{fig:passive_setup_results}(a) shows the setup for an obstacle angle of \changed{$15^{\circ}$}. 

\begin{figure*}[t]
    \centering
    \includegraphics[width=\textwidth]{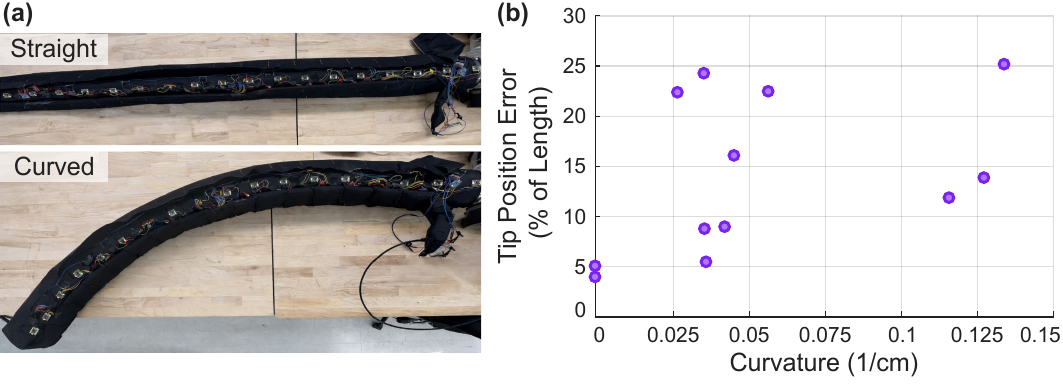}
    \vspace{-2.0em}
    \caption{\textbf{Active steering test setup and results.} 
    \changed{(a) At full length (1.73 m), a single series pouch motor was inflated to various pressures to induce various curvatures while the robot remained on a flat surface.} 
    \changed{(b) Tip position error as a percentage of the robot's length plotted as a function of curvature. Each point represents one trial. No statistically significant relationship between curvature and tip position error was observed ($p = 0.38$). On average, the error is approximately 16\% of the robot length.}
    }
    \label{fig:active_setup_results}
    \vspace{-1.50em}
\end{figure*}

Figure \ref{fig:passive_setup_results}(b) shows \changed{tip position error across the tested obstacle bend angles. No statistically significant relationship between bend angle and tip position error was observed ($p = 0.8$).} The error due to passive bending was typically around $11^{\circ}$ for all bend angles between $0^{\circ}$ and $90^{\circ}$. On average, there was about 84 seconds between offsetting and recording data. 

\subsection{Active Steering Test}
\label{active}

The active steering test (Figure \ref{fig:active_setup_results}) examines how tip position error is affected by robot curvature. The test setup for this experiment involved placing the vine robot on a flat surface, and inflating one series pouch motor through a range of pressures, resulting in various vine curvatures between 0 and 0.15 cm$^{-1}$. The robot's length was kept constant at the fully grown length (1.73 m), and the robot was only steered in the horizontal plane. Between each series pouch motor inflation, we completed an offsetting procedure, as described in Section \ref{testarch}. The test procedure is shown in Figure \ref{fig:active_setup_results}(a).

The tip position error results in Figure \ref{fig:active_setup_results}(b) \changed{do not show a statistically significant relationship between curvature and tip position error ($p = 0.38$). The mean tip position error across trials is approximately 16\% of the robot length.} On average, there was about 87 seconds of drift between offsetting and recording data. 

\subsection{Length Test}

\begin{figure*}[t]
    \centering
    \includegraphics[width=\linewidth]{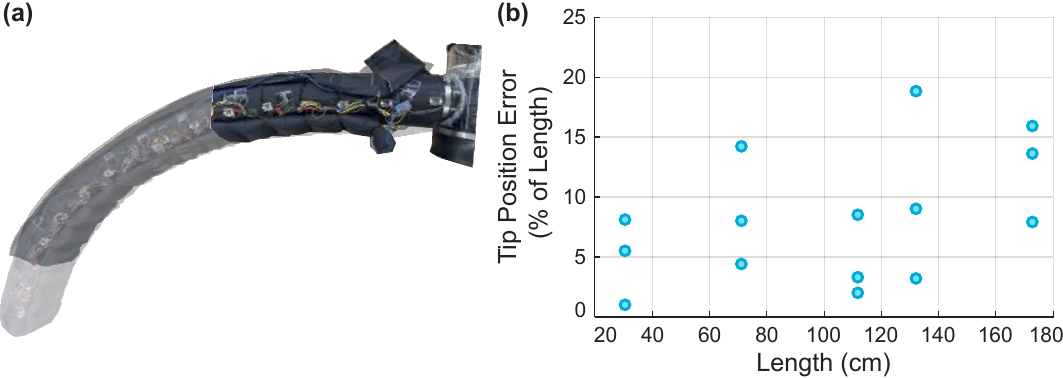}
    \vspace{-2.0em}
    \caption{\textbf{Length test setup and results.} 
    \changed{(a) Vine robot grown to variable lengths while actively steered.
    (b) Tip position error (percentage of robot length) plotted as a function of robot length. Each point represents one trial. A positive trend between length and error was observed, but the relationship did not reach statistical significance ($p$ = 0.10). The average error across trials was approximately $8\%$ of robot length.}
    }
    \label{fig:length_setup_results}
    \vspace{-0.5em}
\end{figure*}

Because shape estimation errors accumulate along the robot body, longer vine robots may exhibit increased tip position error. The goal of this experiment (Figure \ref{fig:length_setup_results}) is to determine the pattern of the error increase as a function of robot length. In this experiment, the vine robot grows continuously and is steered actively, with lengths varied from 30 to 175~cm. The IMU data at selected time points is passed to the shape sensing algorithm, and, using a corresponding photo as the ground truth, the error in tip position is computed for each model. Figure \ref{fig:length_setup_results}(a) demonstrates the experimental setup. Between each length measurement, the IMUs were offset, with an average time of 41 seconds 
between offsetting and recording data. 

\begin{figure*}[ht]
    \centering
    \includegraphics[width=\textwidth]{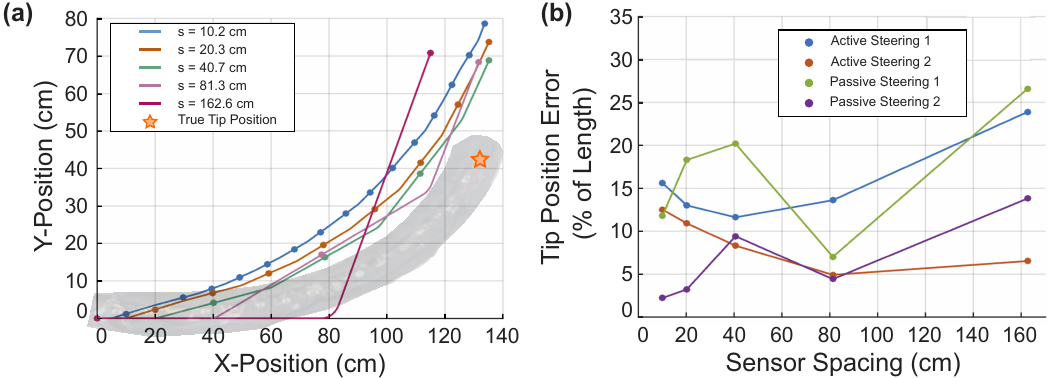}
    \vspace{-2.0em}
    \caption{\textbf{Sensor spacing test setup and results.}
    \changed{IMU data from two passive and two active steering trials were reanalyzed while considering only sensors spaced a distance $s$ apart.
    (a) Estimated robot centerlines for five sensor spacings ranging from 10.2 to 162.6~cm.
    (b) Tip position error (percentage of robot length) as a function of sensor spacing. While one passive trial achieved its lowest error at the smallest spacing, the remaining three trials exhibited minimum error at intermediate spacings (40–80 cm), indicating that denser sensing did not consistently yield lower error.}
    }
    \label{fig:spacing_setup_results}
    \vspace{-1.5em}
\end{figure*}

\changed{The length test results, presented in Figure \ref{fig:length_setup_results}(b), indicate a positive trend between vine robot length and tip position error; however, this relationship did not reach statistical significance ($p = 0.10$)}.

\subsection{Sensor Spacing Test}

This experiment (Figure \ref{fig:spacing_setup_results}) was done by reanalyzing the IMU data from two trials each of the Passive Steering and Active Steering Tests described in the previous subsections, while only considering data from certain IMUs. For example, a trial was completed where only the data from every other IMU was considered, effectively testing the shape sensing accuracy when IMUs were spaced every 20.3 cm. 
The conceptual procedure for this experiment is shown in Figure \ref{fig:spacing_setup_results}(a). The same ground truth photos were used for all sensor spacing trials within one test of interest. During testing, there was an average drift time of 57 seconds between offsetting and recording data. 

The results in Figure \ref{fig:spacing_setup_results}(b) \changed{shows that the optimal sensor spacing was not consistently the smallest tested value. While one passive steering trial achieved its lowest error at 10.2 cm spacing, the other three trials reached minimum error at intermediate spacings (approximately 40–80 cm).}
This is interesting because intuition would lead us to believe that more sensors is always better. However, each additional sensor adds noise and drift to the data, and more IMUs means a higher chance that some are producing significant errors. 

\section{Discussion and Limitations}

\begin{table}[t]
\small
\centering
\caption{Summary of Experimental Findings}
\label{tab:summary}
\setlength{\tabcolsep}{5pt}
\renewcommand{\arraystretch}{1.1}
\begin{tabular}{lcc}
\hline
\textbf{Experiment} & \textbf{Mean Tip Error} & \textbf{Significance} \\
\hline
Passive steering & 11\% of Length & $p=0.8$ \\
Active steering & 16\% of Length  & $p=0.38$ \\
Length test & 8\% of Length & $p=0.10$ \\
Drift rate & 1.33$^{\circ}$/min & $p=0.0024$ \\
Sensor spacing & Intermediate best & --- \\
\hline
\end{tabular}
\end{table}

\changed{In this work, we evaluated the performance of distributed IMU-based shape sensing on an actively steered vine robot, quantifying drift behavior and experimentally characterizing how steering mode, robot length, and sensor spacing influence tip position accuracy. Table~\ref{tab:summary} provides a summary of key quantitative findings of our investigations.}

Drift test results for 15 IMUs indicated that some IMUs drifted dramatically more than others; on average, the IMUs drifted at a rate of $1.33^{\circ}$ per minute. \changed{Although the offsetting procedure limits short-term error accumulation, extended deployments without recalibration may lead to significant orientation error.} Either frequent resetting of the robot shape to a known state or clever matching of the location of IMUs inside and outside the robot would help with shape sensing accuracy over time.

During passive steering, the tip position error was approximately 11\% of the robot's length on average, with a slight positive correlation between tip position error and passive steering bend angle. During active steering, the tip position error was higher, approximately 16\% of the robot's length on average. \changed{For a 1.73~m robot, this corresponds to roughly 28~cm of positional error, which is not negligible for many real-world applications such as navigation through confined rubble environments. This increase in error is consistent with the fact that the geometric model assumes a single-bend behavior that more closely resembles passive steering than actively pressurized actuation. Incorporating additional information about actuator pressures, chamber dynamics, and the known eversion mechanics of the robot, particularly the constraint that the tail remains fixed inside the base while new material everts at the tip, could substantially improve shape estimation accuracy.} 

During active steering at various robot lengths, the tip position error was approximately 8\% of the robot's length on average, with a positive correlation between tip position error as a percentage of length and length. This suggests that beyond a certain length, \changed{accumulated modeling and sensor errors may render the current} shape sensing \changed{approach insufficient for precise navigation tasks. Improving performance at longer lengths will likely require algorithms that explicitly account for, as previously mentioned, eversion constraints and robot kinematics, or the integration of complementary sensing modalities.}


\changed{Analysis of sensor spacing revealed that minimum tip error occurred at intermediate spacings (approximately 40–80 cm for the configurations tested), rather than at the smallest spacing. This suggests a trade-off between geometric resolution and cumulative sensor noise: while denser sensing improves curvature representation, additional IMUs increase the probability of drift or individual sensor error affecting the reconstructed shape. These results indicate that optimal spacing depends on both shape complexity and sensor reliability.}

\changed{An important consideration is how distributed IMU-based shape sensing compares to simpler tip-only approaches such as~\cite{qin20253d}, which use a single IMU at the robot tip together with base length estimation to localize the robot. The mean tip errors observed in this study (8–16\% of robot length) do not clearly improve upon those achievable with lower-complexity methods when evaluating tip accuracy alone, suggesting that for applications focused solely on tip localization, a tip-mounted IMU may offer a simpler and more cost-effective solution. The main benefit of distributed sensing, however, is its ability to sense the body. Unlike tip-only methods, which rely on several geometric assumptions, distributed IMUs directly measure orientation along the backbone and can capture multi-bend or non-planar configurations. At the same time, our sensor-spacing results indicate diminishing returns with high sensor density, implying that fewer strategically placed IMUs may achieve comparable performance with reduced system complexity. A direct experimental comparison between distributed and tip-only configurations would further clarify these trade-offs.}

\subsection{Limitations}
\changed{While distributed IMU-based shape sensing demonstrates feasibility, several limitations should be acknowledged. Experiments were conducted in controlled laboratory settings with primarily planar, single-curvature shapes, and ground truth was obtained from overhead images, introducing measurement uncertainty. The BNO055 IMUs exhibited non-uniform drift behavior, and although offsetting mitigates short-term drift, longer deployments may accumulate significant orientation error. The geometric model assumes a single hinge located midway between adjacent IMUs and inextensible outer-edge behavior, which more closely represents passive steering than actively pressurized actuation and likely contributes to increased error during active steering. The observed tip error may be insufficient for precise navigation tasks in constrained rubble environments. Improving performance will likely require algorithms that explicitly account for eversion mechanics, actuator pressures, and additional sensing modalities.}

\section{Conclusions and Future Work}
\changed{This study demonstrates that distributed IMU-based shape sensing is feasible for vine robots under both passive and active steering, while also revealing key performance limitations related to drift, actuation modeling, and length-dependent error accumulation. Although the achieved accuracy may be sufficient for coarse situational awareness, further improvements are required for precision navigation in field environments.}
Future work should focus on mitigating drift effects, testing the system in narrow spaces and multi-plane environments, and improving mechanical and algorithmic integration. Incorporating actuator pressure sensing, eversion constraints, and complementary sensing modalities may substantially improve performance. Additionally, the mechanical integration of the IMUs with the robot body should be improved so that the robot can be miniaturized and carry a tip-mounted sensor pod. With further research and development, a distributed IMU-based shape sensing system is a promising solution for improving the capabilities of vine robots within rubble sites, which will help save the lives of collapse zone survivors.


\bibliographystyle{IEEEtran}
\bibliography{library}

@Misc{collapsebasics,
   author =       "Michael Daley",
   title =        "Building Collapse Basics",
   year =         "2018",
   note         = {[Online]. Available: https://www.firehouse.com/rescue/technical-rescue/structural-collapse/article/20997907/building-collapse-basics-for-firefighters. [Accessed: 20-February-2026]}
}

@Misc{osha,
   author = "Occupational Safety and Health Administration",
   title = "Structural Collapse Guide",
   note   = {[Online]. Available: https://www.osha.gov/emergency-preparedness/guides/structural-collapse. [Accessed: 20-Feb-2026]},
   year = "",
}

@INPROCEEDINGS{snake,
  author={Whitman, Julian and Zevallos, Nico and Travers, Matt and Choset, Howie},
  booktitle={IEEE International Symposium on Safety, Security, and Rescue Robotics}, 
  title={Snake Robot Urban Search After the 2017 {Mexico City} Earthquake}, 
  year={2018},
  volume={},
  number={},
  pages={1-6},
  doi={10.1109/SSRR.2018.8468633}}

@article{vinebasic,
  title={Vine robots},
  author={Coad, Margaret M and Blumenschein, Laura H and Cutler, Sadie and Zepeda, Javier A Reyna and Naclerio, Nicholas D and El-Hussieny, Haitham and Mehmood, Usman and Ryu, Jee-Hwan and Hawkes, Elliot W and Okamura, Allison M},
  journal={IEEE Robotics \& Automation Magazine},
  volume={27},
  number={3},
  pages={120--132},
  year={2020},
  publisher={IEEE}
}

@INPROCEEDINGS{charm,
  author={Gruebele, Alexander M. and Zerbe, Andrew C. and Coad, Margaret M. and Okamura, Allison M. and Cutkosky, Mark R.},
  booktitle={IEEE International Conference on Soft Robotics}, 
  title={Distributed Sensor Networks Deployed Using Soft Growing Robots}, 
  year={2021},
  volume={},
  number={},
  pages={66-73},
  doi={10.1109/RoboSoft51838.2021.9479345}
}

@INPROCEEDINGS{RoBoa,
  author={der Maur, Pascal Auf and Djambazi, Betim and Haberthür, Yves and Hörmann, Patricia and Kübler, Alexander and Lustenberger, Michael and Sigrist, Samuel and Vigen, Oda and Förster, Julian and Achermann, Florian and Hampp, Elias and Katzschmann, Robert K. and Siegwart, Roland},
  booktitle={IEEE International Conference on Soft Robotics}, 
  title={{RoBoa}: Construction and Evaluation of a Steerable Vine Robot for Search and Rescue Applications}, 
  year={2021},
  volume={},
  number={},
  pages={15-20},
  doi={10.1109/RoboSoft51838.2021.9479192}
}

@INPROCEEDINGS{sprout,
  author={McFarland, Ciera and Dhawan, Ankush and Kumari, Riya and Council, Chad and Coad, Margaret and Hanson, Nathaniel},
  booktitle={IEEE International Symposium on Safety Security Rescue Robotics}, 
  title={Field Insights for Portable Vine Robots in Urban Search and Rescue}, 
  year={2024},
  volume={},
  number={},
  pages={190-197},
  doi={10.1109/SSRR62954.2024.10770046}
}

@Misc{BNO055DS,
   author = "Bosch Sensortec",
   title = "{BNO055} Intelligent 9-axis absolute orientation sensor data sheet",
   note   = {[Online]. Available: https://www.bosch-sensortec.com/products/smart-sensors/bno055/. [Accessed: 20-Feb-2026]},
   pages = {23-47},
   year =  "2014"
}

@article{tape_def,
 author  = {Artem Dementyev and Hsin-Liu (Cindy) Kao and Joseph A. Paradiso},
 title   = {{SensorTape}: Modular and Programmable {3D}-Aware Dense Sensor Network on a Tape},
 journal = {Proceedings of the ACM Symposium on User Interface Software \& Technology},
 year    = {2015},
 pages = {649-658},
 doi = {10.1145/2807442.280750}
}

@article{sheet_shape,
 author  = {Dylan Shah and Stephanie J. Woodman and Lina Sanchez-Botero and Shanliangzi Liu and Rebecca Kramer-Bottiglio},
 title   = {Stretchable Shape-Sensing Sheets},
 journal = {Advanced Intelligent Systems},
 year    = {2023},
 volume = {5},
 number = {12},
 pages = {2300343},
 doi = {10.1002/aisy.202300343}
}

@article{IMU_shape_2,
 author  = {Josie Hughes and Francesco Stella and Cosimo Della Santina and Daniela Rus},
 title   = {Sensing Soft Robot Shape Using {IMUs}: An Experimental Investigation},
 journal = {International Symposium on Experimental Robotics},
 year    = {2021},
 pages = {543-552},
 doi = {10.1007/978-3-030-71151-1_48}
}

@INPROCEEDINGS{IMU_along,
  author={Martin, Yves J. and Bruder, Daniel and Wood, Robert J.},
  booktitle={IEEE/RSJ International Conference on Intelligent Robots and Systems}, 
  title={A Proprioceptive Method for Soft Robots Using Inertial Measurement Units}, 
  year={2022},
  volume={},
  number={},
  pages={9379-9384},
  doi={10.1109/IROS47612.2022.9982185}}

@article{shaevitz2025shape,
  title={Shape Reconstruction using Soft, Stretchable Sensors with Embedded Strain Sensors and Inertial Measurement Units},
  author={Shaevitz, Aiden and Karam, Joseph and Votzke, Callen and Ling, Benny and Batten, Belinda A and Johnston, Matthew L and Davidson, Joseph R},
  journal={IEEE Sensors Journal},
  year={2025},
  volume={25},
  number={19},
  pages={36930 - 36939},
  publisher={IEEE}
}

@article{al2025tip,
  title={Tip-growing robots: Design, theory, application},
  author={Al Harthy, Shamsa and Sadati, SM Hadi and Girerd, C{\'e}dric and Kim, Sukjun and Mondini, Alessio and Wu, Zicong and Saldarriaga, Brandon and Seneci, Carlo A and Mazzolai, Barbara and Morimoto, Tania K and others},
  journal={IEEE Transactions on Robotics},
  year={2025},
  volume={41},
  pages={5511-5532},
  publisher={IEEE}
}

@article{hawkes2017soft,
  title={A soft robot that navigates its environment through growth},
  author={Hawkes, Elliot W and Blumenschein, Laura H and Greer, Joseph D and Okamura, Allison M},
  journal={Science Robotics},
  volume={2},
  number={8},
  pages={eaan3028},
  year={2017},
  publisher={American Association for the Advancement of Science}
}

@incollection{mishima2006development,
  title={Development of pneumatically controlled expandable arm for search in the environment with tight access},
  author={Mishima, Daisuke and Aoki, Takeshi and Hirose, Shigeo},
  booktitle={Field and Service Robotics: Recent Advances in Reserch and Applications},
  pages={509--518},
  year={2006}
}

@article{ataka2020model,
  title={Model-based pose control of inflatable eversion robot with variable stiffness},
  author={Ataka, Ahmad and Abrar, Taqi and Putzu, Fabrizio and Godaba, Hareesh and Althoefer, Kaspar},
  journal={IEEE Robotics and Automation Letters},
  volume={5},
  number={2},
  pages={3398--3405},
  year={2020},
  publisher={IEEE}
}

@INPROCEEDINGS{frias,
  author={Frias-Miranda, Eugenio and Srivastava, Alankriti and Wang, Sicheng and Blumenschein, Laura H.},
  booktitle={IEEE/RSJ International Conference on Intelligent Robots and Systems}, 
  title={Vine Robot Localization Via Collision}, 
  year={2023},
  volume={},
  number={},
  pages={2515-2521},
  doi={10.1109/IROS55552.2023.10342238}}

@article{qin20253d,
  title={{3D} Steering and Localization in Pipes and Burrows using an Externally Steered Soft Growing Robot},
  author={Qin, Yimeng and Grinberg, Jared and Heap, William and Okamura, Allison M},
  journal={arXiv preprint arXiv:2507.07225},
  year={2025}
}

\end{document}